%% file: main.tex
\documentclass[10pt,twocolumn,letterpaper]{article}

\usepackage[accsupp]{axessibility} 
\usepackage{authblk}
\usepackage{cvpr}              
\usepackage{times}
\usepackage{epsfig}
\usepackage{graphics}
\usepackage{graphicx}
\usepackage{amsmath}
\usepackage{amssymb}
\usepackage{booktabs}
\usepackage{amsthm}
\usepackage{multirow}
\usepackage{bbding}
\usepackage[table]{xcolor} 

\definecolor{cvprblue}{rgb}{0.21,0.49,0.74}
\usepackage[pagebackref,breaklinks,colorlinks,citecolor=cvprblue]{hyperref}

\def\paperID{2034}
\def\confName{CVPR}
\def\confYear{2024}

\title{Grounding and Enhancing Grid-based Models for Neural Fields}

\author[1]{Zelin Zhao}
\author[2]{Fenglei Fan\thanks{Corresponding author.}}
\author[1,3]{Wenlong Liao}
\author[1]{Junchi Yan}
\affil[1]{Department of CSE \& MoE Key Lab of AI,  Shanghai Jiao Tong University}
\affil[2]{Department of Mathematics, The Chinese University of Hong Kong}
\affil[3]{Cowa Tech. Ltd.}
\newcommand{\modelname}{MulFAGrid}
\newcommand{\bO}{\boldsymbol{O}}
\newcommand{\bY}{\boldsymbol{Y}}

\newcommand{\bs}{\boldsymbol{s}}

\newcommand{\bX}{\boldsymbol{X}}
\newcommand{\bx}{\boldsymbol{x}}
\newcommand{\bz}{\boldsymbol{z}}
\newcommand{\bZ}{\boldsymbol{Z}}
\newcommand{\br}{\boldsymbol{r}}
\newcommand{\bp}{\boldsymbol{p}}
\newcommand{\bw}{\boldsymbol{w}}
\newcommand{\mg}{\boldsymbol{G}}
\newcommand{\dui}{\CheckmarkBold}

\newcommand{\bep}{\boldsymbol{\varepsilon}}
\newcommand{\bv}{\boldsymbol{v}}
\newcommand{\ml}{\mathcal{L}}
\newcommand{\bi}{\boldsymbol{I}}
\newcommand{\bt}{\boldsymbol{T}}
\newcommand{\budui}{\XSolidBrush}
\definecolor{gray}{HTML}{efefef}
\newcommand{\tablestyle}[2]{\setlength{\tabcolsep}{#1}\renewcommand{\arraystretch}{#2}\centering\footnotesize}
\newtheorem{defi}{Definition}
\newtheorem{theorem}{Theorem}
\newcolumntype{x}[1]{>{\centering\arraybackslash}p{#1}}
\newcolumntype{y}[1]{>{\raggedright\arraybackslash}p{#1}}
\newcolumntype{z}[1]{>{\raggedleft\arraybackslash}p{#1}}

\begin{document}
\maketitle
\input{sec/0_abstract}    
\input{sec/1_intro}
\input{sec/2_related_work}
\input{sec/3_method}

\input{sec/4_experimental_results}
\clearpage
{\small
\bibliographystyle{ieeenat_fullname}
\bibliography{main}
}
\input{sec/X_suppl}

\end{document}

%% file: sec/0_abstract.tex
\begin{abstract}
Many contemporary studies utilize grid-based models for neural field representation, but a systematic analysis of grid-based models is still missing, hindering the improvement of those models. Therefore, this paper introduces a theoretical framework for grid-based models. This framework points out that these models' approximation and generalization behaviors are determined by grid tangent kernels (GTK), which are intrinsic properties of grid-based models. The proposed framework facilitates a consistent and systematic analysis of diverse grid-based models. Furthermore, the introduced framework motivates the development of a novel grid-based model named the Multiplicative Fourier Adaptive Grid (MulFAGrid). The numerical analysis demonstrates that MulFAGrid exhibits a lower generalization bound than its predecessors, indicating its robust generalization performance. Empirical studies reveal that MulFAGrid achieves state-of-the-art performance in various tasks, including 2D image fitting, 3D signed distance field (SDF) reconstruction, and novel view synthesis, demonstrating superior representation ability. The project website is available at~\href{https://sites.google.com/view/cvpr24-2034-submission/home}{this link}.
\end{abstract}

%% file: sec/1_intro.tex
\section{Introduction}
\label{sec:intro}

Neural fields~\cite{nerf, NeuralFields} are coordinate-based networks representing a field, a continuous parameterization representing a physical quantity of an object or a scene. These fields have demonstrated significant success in image regression~\cite{fourierFFN}, view synthesis~\cite{nerf}, and 3D model reconstruction~\cite{neus}. Recent studies find that neural field techniques can be applied to visual computing problems and beyond~\cite{NeuralFields}. Therefore, this field is poised to have a topographic impact on computer vision and machine learning.

Recent empirical studies~\cite{plenoxels, dvgo, NFFB, neuRBF, 3DGS} have substantiated that grid-based models, parameterized by grid feature tensors and operating on grids, can achieve a maximum two-orders-of-magnitude speed-up compared to MLP-based neural fields~\cite{InstantNGP, dvgo, nerf}, all while upholding high-fidelity representation quality. Grid-based models can be either regular (based on regular grids~\cite{dvgo,dvgov2,plenoxels}) or irregular (based on point cloud~\cite{3DGS} or mesh~\cite{neuralbody,neus}). However, to the best of our knowledge, there is no known theory to analyze the learning behaviors of grid-based models systematically. Consequently, the empirical success of grid-based models not only lacks a theoretical foundation but also reveals a deficiency in effective principles for designing enhanced grid-based models.

To ground and enhance grid-based models, we propose a theory inspired by neural tangent kernels~\cite{jacot2018NTK, bietti2019inductiveNTK, wang2022andNTKPINN}. Our theory aims to capture the optimization characteristics and generalization performance of grid-based models. Different from neural tangent kernels (NTKs)~\cite{jacot2018NTK} that study behaviors of MLPs, we target grid-based models and propose tangent kernels for them named grid tangent kernels (GTKs). GTKs are defined as the covariance between model gradients with respect to their parameters at two different input data. They describe precisely how changes in grid-based models' parameters impact their predictions during training. We show that GTKs of grid-based models remain unchanged during training so that a grid-based model behavior can be understood as a linear kernelized model when the kernel used in the GTK remains unchanged. Therefore, the GTK is particularly useful for understanding the connection between grid-based model architectures and their training dynamics. Furthermore, we derive a generalization bound based on Rademacher complexity~\cite{NTKarora2019fineGrained}, which measures how well a trained grid-based model performs on unseen data. This property helps researchers interested in neural fields understand the factors influencing generalization in grid-based models. Our numerical studies reveal that the performance of previous grid-based models can be escalated further by designing better structures toward a suitable GTK spectrum and a better generalization bound.

\begin{figure*}[t]
\begin{center}
\includegraphics[width=1.0\linewidth]{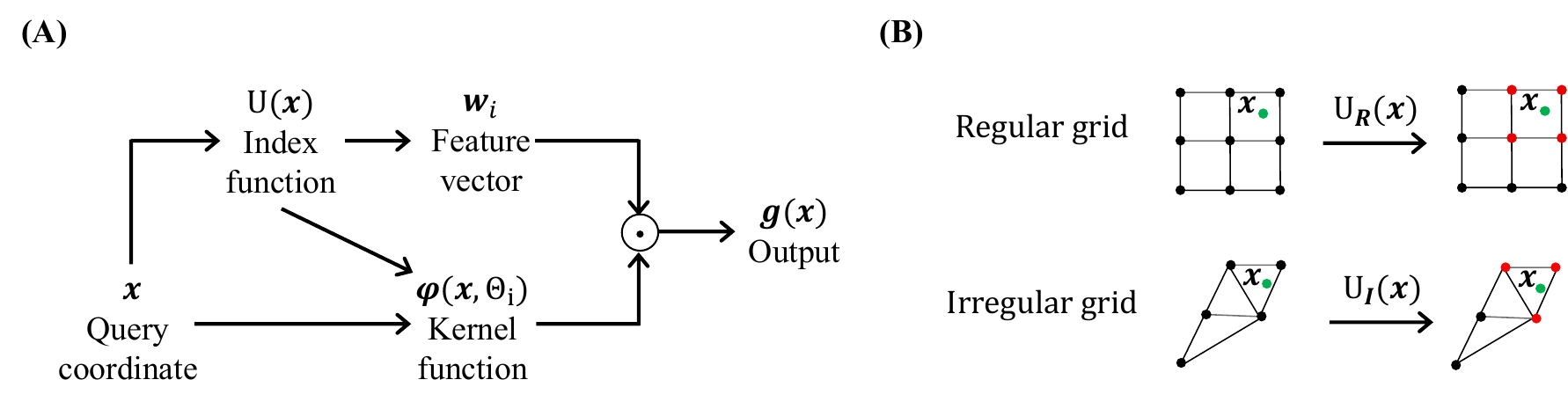}
\end{center}
\vspace{-10pt}
   \caption{Formulations for grid-based models. \textbf{(A)} A grid-based model takes a query coordinate $\bx$ as the input, which is sent to an index function $U$ to acquire a set of feature vectors $\bw$ from the grid. Then, the model outputs a weighted average of the kernel function $\varphi$ and the feature vectors $\bw$.  \textbf{(B)} Our formulation supports grid-based models using a regular grid or an irregular grid, depending on the index function. The query coordinate is shown in green, and the queried points are in red. Please refer to~\cref{sec-grid-model-definition} for more details.}
\label{fig-grid-based-model}
\end{figure*}

\begin{table*}[tb!]
\vspace{-5pt}
    \centering
    \resizebox{\linewidth}{!}{
    \tablestyle{0pt}{1.0}
    \begin{tabular}{x{74pt}| x{54pt} x{54pt} x{45pt} x{54pt} x{50pt} x{64pt} x{54pt} x{60pt}}
\toprule
Method &
  Publication venue &
  3D SDF reconstruction &
  2D image fitting &
  Novel view synthesis &
  Theoretical results &
  Grid-based models &
  Multiplicative filters &
  Support irregular grids \\ \midrule
\midrule
BACON~\cite{bacon}                                  & CVPR22           & \dui   & \dui   & \dui   & \dui   & \budui & \budui & \budui \\
DVGO~\cite{dvgo}                       & CVPR22           & \budui & \budui & \dui   & \budui & \dui   & \budui & \budui           \\
Plenoxels~\cite{plenoxels}             & CVPR22           & \budui & \budui & \dui   & \budui & \dui   & \budui & \budui           \\
MINER~\cite{MINER}                     & ECCV22           & \dui   & \dui   & \dui   & \budui & \budui  & \budui & \budui \\
PNF~\cite{PNF}                                    & NeurIPS22        & \dui   & \dui   & \dui   & \dui   & \budui & \budui & \budui \\
InstantNGP~\cite{InstantNGP}           & SIGGRAPH22       & \dui   & \dui   & \dui   & \budui & \dui   & \budui & \budui           \\
3DGS~\cite{3DGS}     & SIGGRAPH23       & \budui & \budui & \dui   & \budui & \dui   & \budui & \dui             \\
NFFB~\cite{NFFB}                       & ICCV23           & \dui   & \dui   & \dui   & \budui & \dui   & \budui & \budui           \\
NeuRBF~\cite{neuRBF}                   & ICCV23           & \dui   & \dui   & \dui   & \budui & \dui   & \budui & \dui             \\
\rowcolor{gray} \modelname\space(ours) & CVPR24 & \dui   & \dui   & \dui   & \dui   & \dui   & \dui   & \dui   \\ \bottomrule
    \end{tabular}}
    \caption{A detailed technical comparison among other recent neural field models (since 2022) and~\modelname\space.}
    \label{tab-method-comparison}
\end{table*}

Guided by the theory of GTK, we propose a new grid-based model called the Multiplicative Fourier Adaptive Grid (\modelname), which leverages multiplicative filters~\cite{MFN} to model nodal functions with constructed Fourier features, and then the extracted nodal features are normalized via a node-wise normalization function to form kernel functions. Finally, the features are gathered by element-wise multiplication between nodal features and extracted kernel features. An adaptive learning technique is adopted to optimize kernel features and grid features of~\modelname\space jointly. Our model supports both regular and irregular grids, depending on the choice of the index function. The technical comparison between our model and related methods is presented in~\Cref{tab-method-comparison}, where we highlight our method is a general-purpose grid-based model, supporting a wide range of applications such as 2D image fitting~\cite{fourierFFN}, 3D SDF reconstruction~\cite{SIREN} and view synthesis~\cite{nerf}.

We methodologically analyze the characteristics of state-of-the-art grid-based models~\cite{InstantNGP, neuRBF, NFFB} and ours via the proposed GTK, and our mathematical findings indicate that the spectrum of~\modelname's GTK is wider in the high-frequency domain, leading to better learning efficiency in learning high-frequency components. Furthermore, our numerical study in 2D toy examples shows that~\modelname\space has a tighter generalization bound in most regions of the 2D data plane than InstantNGP~\cite{InstantNGP}, NFFB~\cite{NFFB} and NeuRBF~\cite{neuRBF}. Finally, the visualization of image regression experiments verifies the connection between the GTK analysis and predictive performance.

We then conduct systematic experiments on three fundamental tasks utilizing neural fields: 2D image fitting, 3D reconstruction with the signed distance field (SDF), and novel view synthesis via neural radiance fields (NeRF). In 2D image fitting and 3D SDF reconstruction, \modelname\space achieves competitive performance with other grid-based models InstantNGP~\cite{InstantNGP}, NFFB~\cite{NFFB} and NeuRBF~\cite{neuRBF}. In neural radiance fields, we experiment on five benchmarks ranging from bounded scenes and unbounded scenarios, and we find that \modelname\space outperforms previous grid-based models by a notable margin. Moreover, despite the above experiments based on regular grids, we also experiment with non-regular grids (point clouds) and observe that \modelname\space is competitive compared to the strong baseline 3DGS~\cite{3DGS}. However, our rendering speed is much lower than 3DGS~\cite{3DGS}.
\textbf{Our contributions are:}

\begin{enumerate}
  \item We present a theory based on tangent kernels for grid-based models. The proposed theory highlights that the grid tangent kernel (GTK) of grid-based models stays unchanged during training, and the GTK can characterize the generalization bound of grid-based models.
  \item We propose a grid-based model based on multiplicative filters and Fourier features, and we propose an adaptive learning approach to optimize the kernel features and grid features jointly.
  \item We conduct GTK analysis for several state-of-the-art grid-based models, and the numerical studies show that \modelname\space has a better generalization bound than other grid-based models. 
  \item Our empirical results show that \modelname\space achieves state-of-the-art performances in 2D image fitting, 3D SDF reconstruction, and NeRF reconstruction, compared to grid-based models and other methods.
\end{enumerate}

\vspace{-10pt}

%% file: sec/2_related_work.tex
\section{Related work}
\label{sec-related-work}
\textbf{Neural Radiance Fields (NeRF).} Neural radiance fields (NeRF)~\cite{nerf} proposes representing colors and densities via MLPs and learning the implicit 3D representations via differentiable volumetric rendering. After that, NeRF-based approaches dominate novel view synthesis~\cite{mip_nerf, mipnerf360, nerfa, nerfW, guojinPINN}. NeRF has been applied to a wide range of topics, such as generation~\cite{giraffe}, surface reconstruction~\cite{neus}, and SLAM~\cite{nerfslam}. Previous works improve the efficiency of NeRFs in various aspects~\cite{fastnerf, InstantNGP, kilonerf, reiser2023merf}. FastNeRF~\cite{fastnerf} and Instant-NGP~\cite{InstantNGP} apply advanced caching techniques to speed up the training of NeRFs. KiloNeRF~\cite{kilonerf} proposes decomposing the high-capacity MLP into thousands of small MLPs. Grid-based approaches~\cite{plenoxels, dvgo, dvgov2} engage a lot of researchers~\cite{tensorf, wang2022fourierPlenOctrees, blocknerf,liang2022coordx} because they are simple and fast.

\textbf{Unbounded Scene Reconstruction.} When depth cameras or multi-view stereos are available, one can use structure from motion (SfM) to reconstruct 3D scenes~\cite{buildingRome, pmvs, automatedLargeScale,internetMVS,LargeScaleTexturing}. Recent researchers are interested in learning unbounded or large-scale radiance fields~\cite{nerfW, nerf++, mipnerf360, blocknerf,citynerf,urbanRadianceFields,meganerf, switchnerf, nerfW, f2nerf}. Mip-NeRF-360~\cite{mipnerf360} improves the parameterization and efficiency of Mip-NeRF~\cite{mip_nerf}. 3DGS~\cite{3DGS} proposes to represent scenes with 3D Gaussians that achieve strong performance in free-view synthesis.

\textbf{Neural Tangent Kernels (NTKs).} Our analysis is closely related to NTK theories~\cite{jacot2018NTK, bietti2019inductiveNTK, du2019gradientNTK, NTKarora2019fineGrained, lee2019wideNTK}. The neural tangent kernel~\cite{jacot2018NTK, lee2019wideNTK} finds that a wide network of any depth evolves as a linear model under gradient descent. Arora et al.~\cite{NTKarora2019fineGrained} describe the generalization ability by the eigenvalues of
the kernel. The closest work to ours is Matthew et al.~\cite{fourierFFN}, which shows that the Fourier feature mapping helps learn the high-frequency component of low-level representations.

\textbf{Fourier Features.} Random Fourier features (RFF)~\cite{rahimi2007randomFeatures} can accelerate the training of kernel machines and are beneficial to large-scale classification and regression tasks. Fourier neural operators (FNO~\cite{fourierFFN, fnoBound}) demonstrate strong performance in the partial derivative equation (PDE) domain. Fourier feature mappings are critical in the original paper of NeRF~\cite{nerf}. Fourier PlenOctrees~\cite{wang2022fourierPlenOctrees} demonstrates the effectiveness of Fourier feature learning in dynamic scenes. NeuRBF~\cite{neuRBF} proposes a sinusoidal composition technique to fuse features of different frequencies.

\begin{figure*}[t!]
\begin{center}
\includegraphics[width=1.0\linewidth]{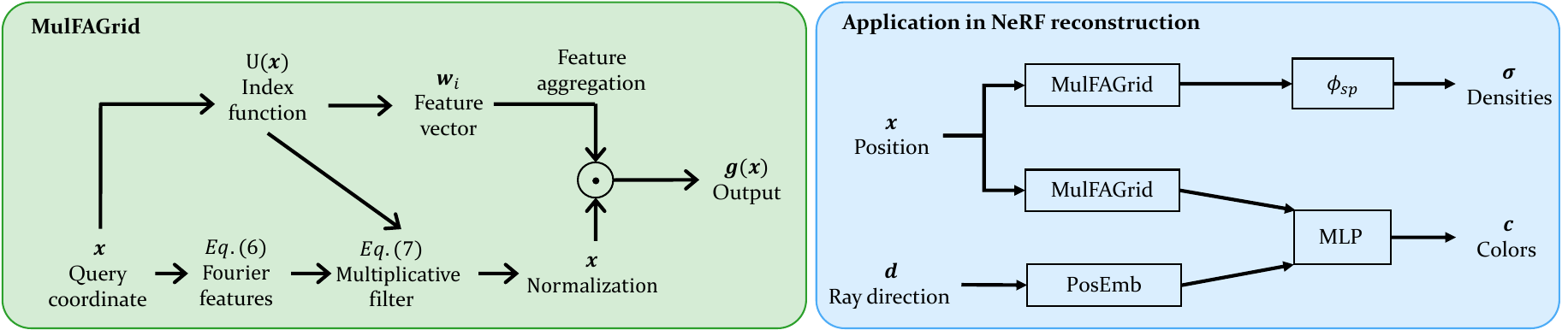}
\end{center}
\vspace{-10pt}
   \caption{(\textbf{Left}) The diagram of~\modelname. The input query coordinate is passed to the multiplicative filter to produce Fourier features and then sent to the normalization layer to compute the aggregation weights. See~\Cref{sec-mulfagrid-model} for details. (\textbf{Right}) The full architecture for neural radiance fields (NeRF). We obtain the densities $\boldsymbol{\sigma}$ via a \modelname\space and the activation $\phi_{sp}$. For the colors $\boldsymbol{c}$, we encode the position $\bx$ via the~\modelname\space with an MLP to post-process the features. After that, we combine the queried spatial features with ray direction information to get color predictions. Please refer to~\Cref{sec-nerf-results} for detailed explanations.}
\label{fig-model-ours}
\vspace{-10pt}
\end{figure*}

%% file: sec/3_method.tex
\section{Methodology}
\subsection{Understanding grid-based models}
\label{sec-GTK}
\subsubsection{Formulations}
\label{sec-grid-model-definition}
Grid-based models play an important role in state-of-the-art neural-field methods because of their efficiency and scalable representation power~\cite{NFFB, neuRBF, 3DGS, dvgo, plenoxels, InstantNGP}. As shown in~\Cref{fig-grid-based-model}, we define a grid-based model $g(\bx, \bw)$ as a machine-learning model with the weighted-average form:
\begin{defi} Given the input query coordinate $\bx$, and $\varphi$ as the kernel function of the grid-based model parameterized by $\Theta$ ($\Theta$ is an empty set if $\varphi$ has no parameters), $\bw_i$ is the weight vector associated with the node $i$, and $U(\bx)$ is an index function which returns a set of indices given the location $\bx$, a grid-based model is defined as a four-element tuple $<\varphi, \Theta, U, \bw>$ with the following computation:
\begin{equation}
g(\bx, \bw)=\sum_{i \in U(\bx)} \varphi\left(\bx, \Theta_i \right) \bw_i.
\label{eq-grid-model}
\end{equation}
\end{defi}
For example, in the simplest case of regular grids, $U(\bx)$ returns the eight nearest grid points around $\bx$, $\varphi$ is the bilinear interpolation with interpolation weights $\Theta$, and $\bw$ denotes features stored in grid points. As shown in~\Cref{fig-grid-based-model} \textbf{(B)}, the formulation in~\Cref{eq-grid-model} supports both \textit{regular grid-based models and irregular grid-based models}, where the main difference between them is the index function $U$. The regular grid-based model uniformly discretizes the coordinate domain with equal intervals, which makes the index function fast. On the other hand, the irregular grid-based model leverages the geometric prior from point-cloud~\cite{3DGS} or meshes~\cite{neus} and does not regularly discretize the coordinate domain. 
\vspace{-8pt}
\subsubsection{The grid tangent kernel (GTK) theory}

To analyze existing grid-based models~\cite{dvgo, plenoxels, NeuralFields, NFFB, neuRBF} for neural fields, we propose a theory for grid-based models inspired by the neural tangent kernel (NTK)~\cite{jacot2018NTK}. Based on the formulation of grid-based models, we introduce a new term called \textit{Grid Tangent Kernel (GTK)} to denote the tangent kernel of a grid-based model, which is crucial in understanding training and generalization performance of grid-based models.

\begin{defi}
Let $\bX$ be a collection of input data where $\bX_i$ is the $i$-th data, and $\bw(t)$ is the weight at the training time $t$. The Grid Tangent Kernel (GTK) of the grid-based model $g$ is defined as an $n\times n$ positive semidefinite matrix $\mg_g(t)$ whose (i, j)-th element is:
 \begin{equation}
 [\mg_g(t)]_{i,j} =  \left\langle\frac{\partial g(\bX_i, \bw(t))}{\partial \boldsymbol{w}}, \frac{\partial g(\bX_j, \bw(t))}{\partial \boldsymbol{w}}\right\rangle.
 \label{eq-gtk-definition}
 \end{equation}
\label{def-gtk}
\vspace{-5pt}
\end{defi}

With the GTK, we further introduce some theoretical results based on the GTK, while we put the proofs to those results in the Supplementary~\Cref{supple-sec-proof-theorems} due to space constraints. Our analysis follows a supervised regression setup~\cite{fourierFFN, jacot2018NTK}, where the task is to regress target labels $\bY$ given the input data $\bX$. Firstly, we introduce the following theorem characterizing the training dynamics:

\begin{theorem}
 Let $\bO(t)=(g(\bX_i, \boldsymbol{w}(t)))_{1\leq i \leq n}$ be the outputs of a grid-based model $g$ where $\bX=(\bX_i)_{1\leq i \leq n}$ is the input data at time $t$, and $\bY=(\bY_i)_{1\leq i \leq n}$ is the corresponding label. 
Then $\bO(t)$ follows this evolution:
\begin{equation}
\frac{d \bO(t)}{d t}=-\mg_g(t) \cdot(\bO(t)-\bY).
\label{eq-gtk-dynamics}
\end{equation}
\label{theorem-law}
\vspace{-5pt}
\end{theorem}
According to the above theorem, GTK reflects model training dynamics. Therefore, analyzing GTK is critical to understanding the training behaviors of the grid-based model. Beyond this result, we further uncover a conservation property of the GTK of grid-based models:
\begin{theorem}
The GTK of a grid-based model $g$, denoted by $\mg_{g}$, stays stationary during training. Formally, this property can be written as:
\begin{equation}
\mg_{g}(t) = \mg_{g}(0),
\end{equation}
where $\mg_{g}(0)$ is the initial GTK of the grid-based model. This property holds for any size of the grid-based model.
\label{theorem-gtk-unchanged}
\end{theorem}

The above theorem shows that the GTK is a fundamental property of the grid-based model, which does not evolve through time. This means the evolution of the output $\bO(t)$ is characterized by~\Cref{eq-gtk-dynamics}, which is an ordinary differentiable equation (ODE). Therefore, grid-based models can be understood as simple linear models~\cite{NTKarora2019fineGrained}. Besides training performance, another important property of a machine-learning model is the generalization gap. We further derive the following generalization bound based on the Rademacher complexity~\cite{NTKarora2019fineGrained}:

\begin{theorem}
Given a probability $\delta_p \in (0, 1)$, suppose the dataset $S=(\bX, \bY)$ contains $n$ i.i.d. samples from a distribution where $n\gg\log\frac{2}{\delta_p}$ and the minimum eigenvalue of the GTK, denoted by $\mg$, is at least a constant $\lambda_0$: $\lambda_{min}(\mg) \geq \lambda_0$. For any grid-based model $g$ that is optimized by gradient descent with a learning rate $\eta_l$, and for any loss function $\mathcal{L}: \mathbb{R}\times\mathbb{R} \rightarrow [0, 1]$, which is 1-Lipschitz in the first argument, we define the population loss as $\mathcal{L}_\mathcal{D}\left(t\right)=\mathbb{E}_{(\bX_i, \bY_i) \sim \mathcal{D}}\left[\mathcal{L}\left(g\left(\bX_i, \bw(t)\right), \bY_i\right)\right]$. Then, with probability at least $1-\delta_p$, a randomly initialized grid-based model trained by gradient descent for $t\geq \Omega\left(\frac{1}{\eta_l \lambda_0} \log \frac{n}{\delta_p}\right)$ iterations has a generalization bound:
\begin{equation}
L_{\mathcal{D}}\left(t\right) \leq \sqrt{\frac{2\bY^{\top}\mg^{-1} \bY}{n}}+O\left(\sqrt{\frac{\log \frac{2}{\delta_p}}{n}}\right).
\label{eq-generalization-bound}
\end{equation}
\label{theorem-gtk-generalization}
\vspace{-15pt}
\end{theorem}
From this theorem, we can find out that the dominating term affecting generalization for a given dataset with a fixed number of data $n$ is $\Delta=\bY^{\top}\mg^{-1} \bY$. If $\Delta$ is tighter, the generalization gap between training and testing data is smaller. This theorem provides another useful tool to assess grid-based models.

\subsection{\modelname}
\label{sec-mulfagrid-model}

\begin{figure}[t!]
\begin{center}
\includegraphics[width=1.0\columnwidth]{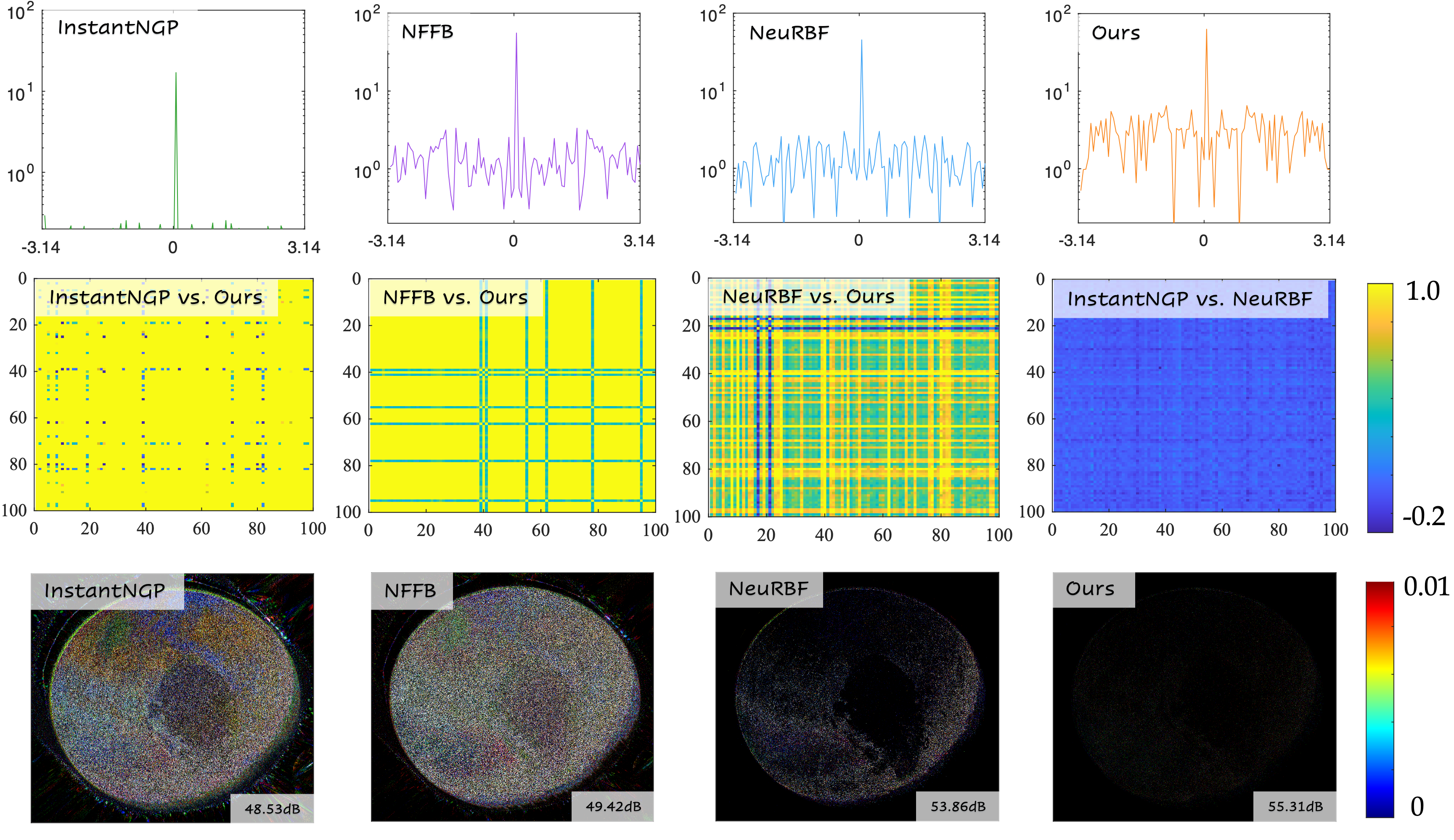}
\end{center}
\vspace{-10pt}
\caption{Analysis of grid-based models (InstantNGP~\cite{InstantNGP}, NFFB~\cite{NFFB}, NeuRBF~\cite{neuRBF}, and ours) based on grid tangent kernels (GTKs) and image regression results. (\textbf{Top}) Visualizations of the GTK Fourier spectrum. \modelname\space has a wide spectrum, especially in the high-frequency domain, leading to faster convergence for high-frequency components~\cite{fourierFFN}. (\textbf{Mid}) Comparisons between generalization bounds of pairs of methods. In this experiment, we construct a dataset, which only contains two data points with labels $\bY=(\bY_1, \bY_2)$, shown in the x-axis and y-axis correspondingly. MulFAGrid has a tighter (lower) generalization bound for most values of $\bY_1$ and $\bY_2$. These findings help explain why MulFAGrid demonstrates better representation ability than other grid-based models. (\textbf{Bot}) Error maps of the fitted images in comparison with ground truth ones.}
\label{fig-gtk-analysis-results}
\vspace{-10pt}
\end{figure}

We propose a new grid-based model called multiplicative Fourier adaptive grid (\modelname), where our numerical study in~\Cref{sec-numerical-study-based-gtk} suggests that it has a faster convergence speed and a tighter GTK-based generalization bound than state-of-the-art grid-based models. Our model is called an adaptive grid-based model because the kernel function $\varphi$ is learned through data. We illustrate the design of~\modelname\space in~\Cref{fig-model-ours}.

\textbf{Fourier features.} We construct a set of Fourier features $\bz^{(1)}$~\cite{nerf, fourierFFN} based on the input query location $\bx$:
\begin{subequations}
\begin{equation}
\gamma(\bx, j) =
    \begin{cases}
        \sin(2^{\lfloor j / 2 \rfloor}\pi \bx), & \text{if } j \bmod 2 = 0,\\
        \cos(2^{\lfloor j / 2 \rfloor}\pi \bx), & \text{else.}
    \end{cases}
\label{eq-fourier-features-1}
\end{equation}
\begin{equation}
\bz^{(1)} = \gamma(\bx, 1)  \oplus \gamma(\bx, 2)  \oplus \ldots  \oplus \gamma(\bx, d_f),
\label{eq-fourier-features-2}
\end{equation}
\label{eq-fourier-features}
\end{subequations}
where $d_f$ is a hyperparameter of the dimension of the constructed Fourier features, and $\oplus$ denotes the concatenation operator. Our theory indicates that the input Fourier features are important in narrowing the generalization bound.

\textbf{Multiplicative filters.} We adopt multiplicative filters~\cite{MFN} to apply non-linearity to the constructed Fourier features and inform the model with the node index $i\in U(\bx)$. The output of multiplicative filters is denoted by $\tilde{\varphi}(\bx, \Theta)$:
\begin{equation}
\begin{aligned}
\bs(i; \theta^{(k)}) &= \sin \left(\omega^{(k)} i+\phi^{(k)}\right),\\
\bz^{(k+1)} & =\left(W^{(k)} \bz^{(k)}+b^{(k)}\right) \circ \bs\left(i; \theta^{(k+1)}\right), \\
\tilde{\varphi}(\bx, \Theta_i) & =W^{(n_m)} \bz^{(n_m)}+b^{(n_m)}.
\end{aligned}
\end{equation}
Here, $k = 1, 2, ..., n_m-1$ is the index of multiplicative filters, $n_m$ is the total number of filters, $\bs$ is the applied sinusoidal filter, $\Theta_i = \left\{\omega^{(k)}, \phi^{(k)}, W^{(k)}, b^{(k)}\right\}$ are parameters in the filters, and $\circ$ denotes the element-wise multiplication.

\textbf{Normalization and feature aggregation.} We apply a node-wise normalization layer to acquire the final kernel:
\begin{equation}
\varphi\left(\bx, \Theta_i\right)=\frac{\tilde{\varphi}\left(\bx, \Theta_i\right)}{\sum_{i \in U(\bx)} \tilde{\varphi}\left(\bx, \Theta_i\right)}.
\label{eq-normalization}
\end{equation}
With this kernel function, we produce the final representation of grid-based models via a feature aggregation procedure according to~\Cref{eq-grid-model}.

\textbf{Adaptive learning.} We propose an adaptive learning method to adapt the kernel function to the data. In theory, kernel parameters $\Theta$ and features of the grid-based model $\bw$ can be iteratively optimized, similar to the expectation-maximization algorithm~\cite{moon1996expectationEM, neuRBF}. Learning of the kernel parameters $\Theta$ contributes to a better GTK, which further narrows the generalization bound. The feature learning stage finds the best weight $\bw$ while the GTK is unchanged during this procedure (according to~\Cref{theorem-gtk-unchanged}). While in experiments, we jointly optimize $\Theta$ and $\bw$ together for simplicity, and we find this work well in practice (see~\Cref{sec-ablation-study}).

%% file: sec/4_experimental_results.tex
\section{Experimental results}
\label{sec-exp-results}
We first provide a numerical study based on the grid tangent kernel (GTK) to analyze the performance of several grid-based models and ours. Then, we evaluate our method and baselines on various applications using neural fields. For each application, we first describe the dataset and baselines and then provide qualitative and quantitative results. Choices of baselines for each task follow previous works~\cite{NFFB, neuRBF}, and we wish to note that some baselines do not support all applications (see~\Cref{tab-method-comparison}).

\subsection{Numerical study based on the GTK}
\label{sec-numerical-study-based-gtk}
In this section, we provide various studies based on the GTK to analyze various grid-based models.

\textbf{Baselines and data.} We compare ours against several grid-based models: InstantNGP~\cite{InstantNGP}, NFFB~\cite{NFFB}, and NeuRBF~\cite{neuRBF}. Each grid-based model is trained until convergence, and we ensure our model does not contain more parameters than baselines. The numerical analysis is conducted in a 2D image dataset composed of the ultra-high-resolution image ``Pluto"~\cite{neuRBF}.



\begin{figure}[t!]
\begin{center} 
\includegraphics[width=1.0\linewidth]{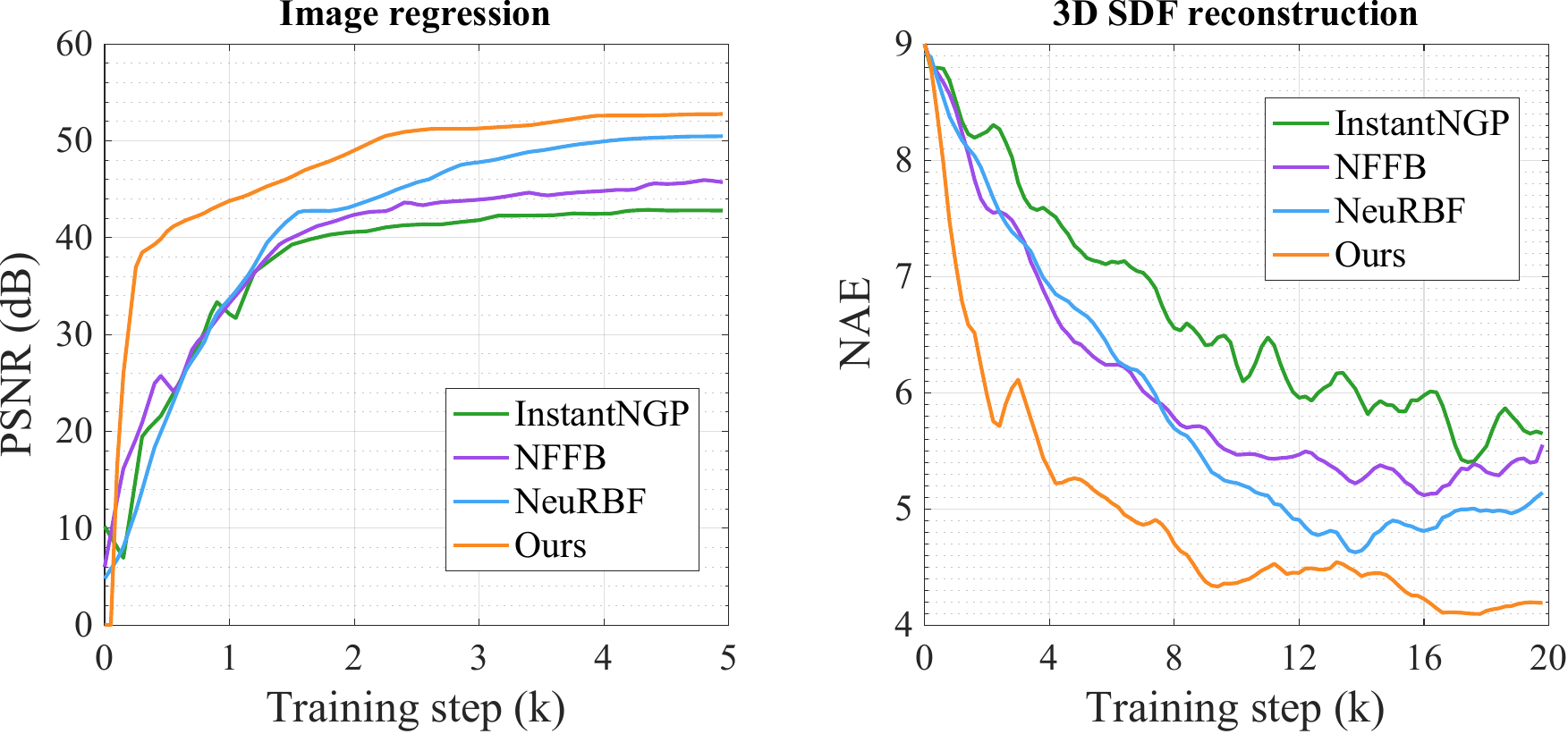}
\end{center}
\vspace{-10pt}
\caption{Comparison curves of several grid-based models: InstantNGP~\cite{InstantNGP}, NFFB~\cite{NFFB}, NeuRBF~\cite{neuRBF}, and~\modelname. (\textbf{Left}) Training curves of the image regression task on the Kodak dataset~\cite{franzen1999kodak}. (\textbf{Right}) The evolution of the normal angular error (NAE) through training of the 3D SDF reconstruction task~\cite{neuRBF}.}
\label{fig-comparison-curves}
\end{figure}

\begin{table}[t!]
\begin{center}
\begin{tabular}{l|llll}
\toprule
           & Steps & Time$\downarrow$   & \# Params$\downarrow$ & PSNR$\uparrow$  \\ \midrule
BACON~\cite{bacon}      & 5k    & 85.2s  & 268K      & 38.51 \\
PNF~\cite{PNF}        & 5k    & 480.9s & 287K      & 38.99 \\ 
InstantNGP~\cite{InstantNGP} & 35k   & 1.9m   & 511K      & 39.14 \\ 
MINER~\cite{MINER}      & 35k   & 14.2m  & 415K      & 39.25 \\ 
NFFB~\cite{NFFB}       & 5k    & 39.4s  & 154K      & 45.28 \\ 
NeuRBF~\cite{neuRBF}     & 5k    & \textbf{28.5s}  & 128K      & 54.84 \\
\rowcolor{gray}Ours       & 5k    & 29.4s  & \textbf{119K}      & \textbf{56.19} \\ \bottomrule
\end{tabular}
\end{center}
\vspace{-10pt}
\caption{2D image fitting results on the validation set of DIV2K dataset~\cite{DIV2K}. Images are center cropped and down-sampled to $256\times256\times3$ following the practice of BACON~\cite{bacon} and NeuRBF~\cite{neuRBF}.}
\label{tab-image-fit}
\end{table}

\begin{table}[t!]
\begin{center}
\begin{tabular}{l|llll}
\toprule
           & Steps & \# Params$\downarrow$ & IoU$\uparrow$ & NAE$\downarrow$ \\ \midrule
NGLOD~\cite{nglod}      & 245k  & 78.84M    & 0.9963        & 6.14            \\ 
InstantNGP~\cite{InstantNGP} & 20k   & 950K      & 0.9994        & 5.70            \\ 
NFFB~\cite{NFFB}       & 20k   & 1.4M      & 0.9994        & 5.23            \\
NeuRBF~\cite{neuRBF}     & 20k   & 856K      & \textbf{0.9995}        & 4.93            \\
\rowcolor{gray}Ours       & 5k    & \textbf{823K}      & \textbf{0.9995}        & \textbf{4.51}            \\ \bottomrule
\end{tabular}
\end{center}
\vspace{-10pt}
\caption{3D signed distance field (SDF) reconstruction results on the sampled 3D models dataset~\cite{neuRBF}.}
\label{tab-3d-recon}
\vspace{-10pt}
\end{table}

\textbf{Spectrum analysis.} ~\Cref{theorem-gtk-unchanged} indicates that the GTK reflects a property of a grid-based model, which makes it possible to understand the training of grid-based models in functional space instead of parametric space~\cite{NTKarora2019fineGrained}. Studies in the NTK domain~\cite{NTKarora2019fineGrained, fourierFFN} show that the spectrum of the NTK reflects their convergence and generalization performance corresponding to different frequencies. Similarly, we conduct a spectrum analysis~\cite{fourierFFN, du2019gradientNTK} of grid-based models. We sample 100 data points from the training image to compute GTKs of different methods, where the ground-truth is the color value, and the calculation follows the GTK's definition~\Cref{eq-gtk-definition}. The value of GTK is normalized to the $[0, 1]$ range, and the Fourier spectrum of the GTK is shown in~\Cref{fig-gtk-analysis-results}. We find that InstantNGP~\cite{InstantNGP} has a peak spectrum at the low frequency, corresponding to a very ``narrow" kernel~\cite{jacot2018NTK}. The drawback of a too-narrow kernel is overfitting to the training points~\cite{fourierFFN}. Recent grid-based models such as NFFB~\cite{NFFB} and NeuRBF~\cite{neuRBF} can mitigate such artifacts by introducing hierarchical structures~\cite{NFFB} or radial basis functions~\cite{neuRBF}. Besides, we find that MulFAGrid has a wider GTK with higher values in high-frequency regions so that it can learn better details. Therefore, MulFAGrid can balance better between ``underfitting" and ``overfitting" extremes~\cite{roberts2022principlesDeepLearning}.

\begin{figure*}[t!]
\begin{center}
\includegraphics[width=1.0\linewidth]{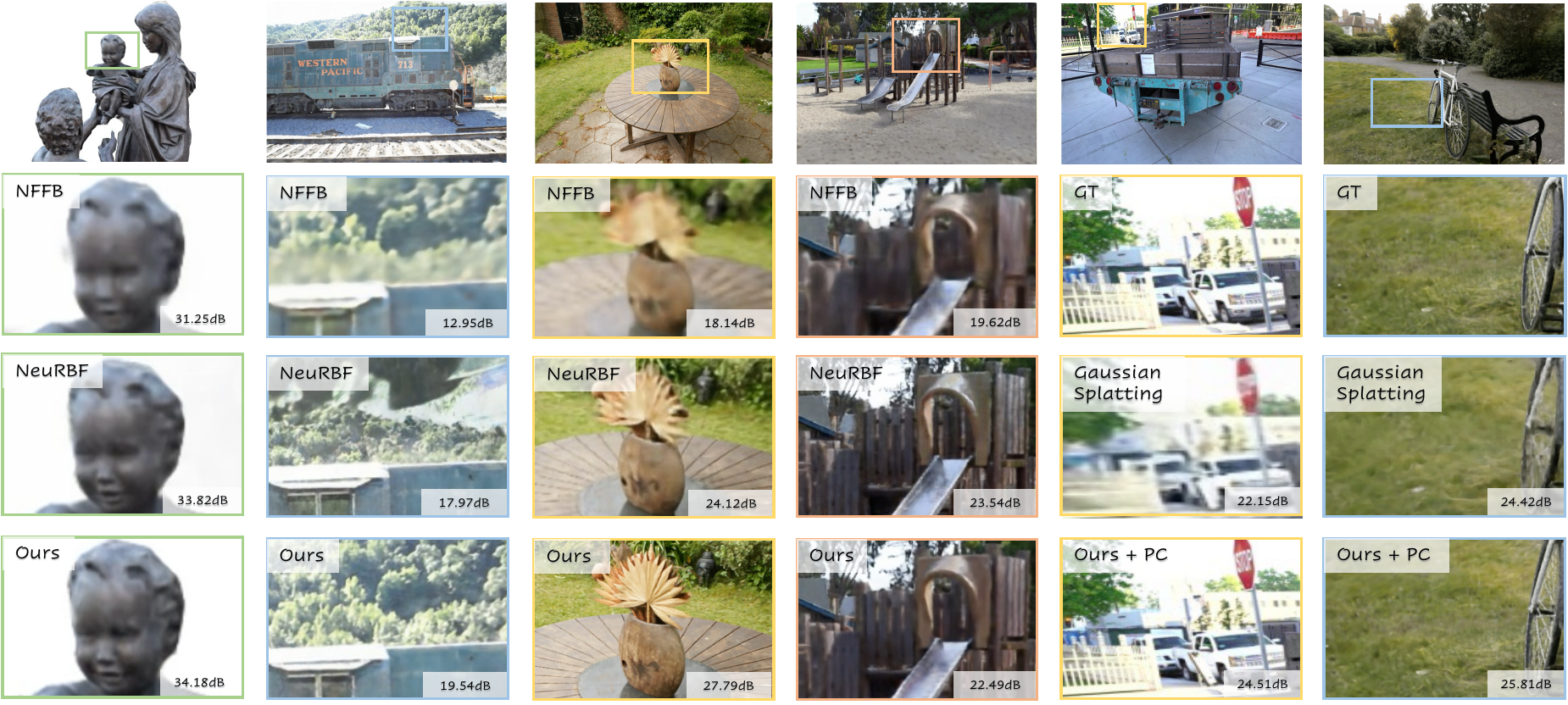}
\end{center}
\vspace{-10pt}
\caption{Rendering results on various scenes from SyntheticNeRF~\cite{nerf}, Tanks\&Temples~\cite{tanksAndTemples} and Mip-NeRF-360~\cite{mipnerf360}. We visualize comparison results against some grid-based models NFFB~\cite{NFFB}, NeuRBF~\cite{neuRBF}, and the strong baseline 3DGS~\cite{3DGS} based on point cloud initialized from structure-from-motion (SfM).}
\label{fig-nerf-visualization-comparison}
\end{figure*}

\begin{table*}[t!]
\centering
\resizebox{\textwidth}{!}{
\begin{tabular}{c|ccc|ccc|ccc|ccc|ccc}
\toprule
Benchmark         & \multicolumn{3}{c|}{SyntheticNeRF~\cite{nerf}}                                          & \multicolumn{3}{c|}{LLFF~\cite{localLightFieldFusion}}                                                   & \multicolumn{3}{c|}{Tanks\&Temples~\cite{tanksAndTemples}}                                         & \multicolumn{3}{c|}{Mip-NeRF-360~\cite{mipnerf360}}                                           & \multicolumn{3}{c}{SFMB~\cite{blocknerf}}                                                    \\ \midrule
Method $\vert$ Metric   & \multicolumn{1}{c|}{Time}  & \multicolumn{1}{c|}{\#Params} & PSNR           & \multicolumn{1}{c|}{Time}  & \multicolumn{1}{c|}{\#Params} & PSNR           & \multicolumn{1}{c|}{Time}  & \multicolumn{1}{c|}{\#Params} & PSNR           & \multicolumn{1}{c|}{Time}  & \multicolumn{1}{c|}{\#Params} & PSNR           & \multicolumn{1}{c|}{Time}   & \multicolumn{1}{c|}{\#Params} & PSNR           \\ \midrule
InstantNGP~\cite{InstantNGP}        & \multicolumn{1}{c|}{\textbf{3.80m}}  & \multicolumn{1}{c|}{12.2M}    & 32.08          & \multicolumn{1}{c|}{26.7m} & \multicolumn{1}{c|}{13.2M}    & 25.28          & \multicolumn{1}{c|}{\textbf{12.4m}} & \multicolumn{1}{c|}{11.3M}    & 19.45          & \multicolumn{1}{c|}{\textbf{20.1m}} & \multicolumn{1}{c|}{12.4M}    & 23.14          & \multicolumn{1}{c|}{\textbf{32.5m}} & \multicolumn{1}{c|}{10.9M}    & 25.52          \\ \midrule
NFFB~\cite{NFFB}              & \multicolumn{1}{c|}{36.9m} & \multicolumn{1}{c|}{18.5M}    & 32.04          & \multicolumn{1}{c|}{35.2m} & \multicolumn{1}{c|}{16.3M}    & 21.25          & \multicolumn{1}{c|}{20.1m} & \multicolumn{1}{c|}{15.9M}    & 18.99          & \multicolumn{1}{c|}{25.4m} & \multicolumn{1}{c|}{19.4M}    & 25.21          & \multicolumn{1}{c|}{35.2m}  & \multicolumn{1}{c|}{16.2M}    & 25.11          \\ \midrule
NeuRBF~\cite{neuRBF}            & \multicolumn{1}{c|}{33.6m} & \multicolumn{1}{c|}{17.7M}   & 34.62          & \multicolumn{1}{c|}{31.1m} & \multicolumn{1}{c|}{18.7M}    & 27.05          & \multicolumn{1}{c|}{19.6m} & \multicolumn{1}{c|}{16.0M}    & 20.12          & \multicolumn{1}{c|}{21.4m} & \multicolumn{1}{c|}{15.2M}    & 26.12          & \multicolumn{1}{c|}{40.1m}  & \multicolumn{1}{c|}{9.24M}     & 25.21          \\ \midrule
DVGOv2~\cite{dvgov2}            & \multicolumn{1}{c|}{10.9m} & \multicolumn{1}{c|}{\textbf{5.20M}}     & 32.80          & \multicolumn{1}{c|}{\textbf{10.9m}} & \multicolumn{1}{c|}{6.90M}     & 26.34          & \multicolumn{1}{c|}{21.4m} & \multicolumn{1}{c|}{5.70M}     & 20.10          & \multicolumn{1}{c|}{20.3m} & \multicolumn{1}{c|}{5.64M}     & 25.42          & \multicolumn{1}{c|}{37.8m}  & \multicolumn{1}{c|}{7.68M}     & 26.42          \\ \midrule
Plenoxels~\cite{plenoxels}         & \multicolumn{1}{c|}{11.4m} & \multicolumn{1}{c|}{194M}   & 31.71          & \multicolumn{1}{c|}{24.5m} & \multicolumn{1}{c|}{500M}     & 26.29          & \multicolumn{1}{c|}{20.8m} & \multicolumn{1}{c|}{523M}     & 20.40          & \multicolumn{1}{c|}{21.9m} & \multicolumn{1}{c|}{511M}     & 20.59          & \multicolumn{1}{c|}{39.4m}  & \multicolumn{1}{c|}{524M}     & 25.98          \\ \midrule
\rowcolor{gray}Ours              & \multicolumn{1}{c|}{10.3m} & \multicolumn{1}{c|}{6.14M}     & \textbf{34.68} & \multicolumn{1}{c|}{24.2m} & \multicolumn{1}{c|}{\textbf{5.10M}}     & \textbf{27.22} & \multicolumn{1}{c|}{19.2m} & \multicolumn{1}{c|}{\textbf{5.32M}}     & \textbf{20.85} & \multicolumn{1}{c|}{20.4m} & \multicolumn{1}{c|}{\textbf{5.43M}}     & \textbf{28.98} & \multicolumn{1}{c|}{33.1m}  & \multicolumn{1}{c|}{\textbf{7.21M}}     & \textbf{29.15} \\ \midrule \midrule
3DGS~\cite{3DGS} & \multicolumn{1}{c|}{\textbf{10.7m}} & \multicolumn{1}{c|}{14.9M}    & 33.15          & \multicolumn{1}{c|}{26.5m} & \multicolumn{1}{c|}{\textbf{15.2M}}    & 27.25          & \multicolumn{1}{c|}{27.0m} & \multicolumn{1}{c|}{\textbf{15.3M}}    & 22.22          & \multicolumn{1}{c|}{\textbf{41.2m}} & \multicolumn{1}{c|}{\textbf{15.9M}}    & 27.21          & \multicolumn{1}{c|}{\textbf{35.1m}}  & \multicolumn{1}{c|}{\textbf{7.34M}}     & 29.44          \\ \midrule
\rowcolor{gray}Ours (w/ PC)      & \multicolumn{1}{c|}{10.9m} & \multicolumn{1}{c|}{\textbf{13.3M}}     & \textbf{34.69} & \multicolumn{1}{c|}{\textbf{26.2m}} & \multicolumn{1}{c|}{16.1M}    & \textbf{27.50} & \multicolumn{1}{c|}{\textbf{26.5m}} & \multicolumn{1}{c|}{16.2M}    & \textbf{22.45} & \multicolumn{1}{c|}{42.1m} & \multicolumn{1}{c|}{16.5M}    & \textbf{29.11} & \multicolumn{1}{c|}{36.1m}  & \multicolumn{1}{c|}{\textbf{7.34M}}     & \textbf{30.12} \\ \bottomrule
\end{tabular}
}
\vspace{-5pt}
\caption{Comparison results on five NeRF benchmarks. We report training time, the number of parameters, and PSNR. Reported numbers are averaged across all scenes in each benchmark. The bottom two models are based on point cloud initializations, while the rest do not use such information.}
\label{tab-nerf-large-table}
\vspace{-10pt}
\end{table*}

\textbf{Generalization bound comparisons.} In~\Cref{theorem-gtk-generalization}, we show a theoretical result that provides an upper bound on the generalization error of a grid-based model. The GTK-based generalization bound can measure the generalization ability of a grid-based model without actually training it. As shown in~\Cref{eq-generalization-bound}, the dominating term affecting generalization for a given dataset with a fixed number of data is $\Delta=\bY^{\top}\mg^{-1} \bY$. We visualize the difference in generalization bounds between two grid-based models. Each plot characterizes the generalization bound difference $\Delta_A - \Delta_B$ between two grid-based models, \textit{A} and \textit{B}. We consider a two-point dataset where the ground true values of this dataset are $\bY = (\bY_1, \bY_2)$. We consider all possible values of the vector $\bY$ and visualize the generalization bound difference $\Delta_A - \Delta_B$ in a 2D plane, where $\Delta_A - \Delta_B > 0$ means the generalization bound of the model $B$ is better than that of model $A$. The result is shown in the middle of~\Cref{fig-gtk-analysis-results}. We observe that~\modelname\space has tighter generalization bounds than InstantNGP~\cite{InstantNGP} and NFFB~\cite{NFFB} in almost all possible data combinations. Meanwhile, the generalization bound of~\modelname\space is tighter than NeuRBF~\cite{neuRBF} in most regions. These facts indicate that \modelname\space has better generalization performance than the compared grid-based models. We further show the image regression results at the bottom of~\Cref{fig-gtk-analysis-results}, which verifies that \modelname\space has quicker optimization performance in modeling 2D neural fields than other grid-based models.

\subsection{2D image fitting}
We then evaluate the task of fitting large-scale 2D images. The task goal is to fit an image based on spatial coordinates. The model is trained via the mean squared error, and all grid-based methods are based on regular grids in this task.

\textbf{Baselines.} Despite the mentioned grid-based models before (InstantNGP~\cite{InstantNGP}, NFFB~\cite{NFFB} and NeuRBF~\cite{neuRBF}), we also compare to other recent works MINER~\cite{MINER}, BACON~\cite{bacon} and PNF~\cite{PNF}.

\textbf{Dataset.} Following previous work~\cite{neuRBF}, we use the \texttt{val} split of the DIV2K dataset~\cite{DIV2K} consisting of 100 natural images of 2K resolution with a diversity of contents. Besides, we incorporate the Kodak dataset~\cite{franzen1999kodak} whose images are sampled by the previous work NeuRBF~\cite{neuRBF}.

\textbf{Results.} We present the learning curve of 2D image fitting, comparing to other grid-based models, in~\Cref{fig-comparison-curves}. We observe that our method reaches a high PSNR at a faster speed than the state-of-the-art NeuRBF~\cite{neuRBF}. The quantitative results comparing a wide range of baselines are presented in~\Cref{tab-image-fit}. Our model achieves a better PSNR with fewer parameters, demonstrating the ability to represent target signals precisely.

\subsection{3D signed distance fields reconstruction}
We further validate our framework in reconstructing 3D models represented in 3D signed distance fields (SDF). In this application, all grid-based models use regular grids.

\textbf{Baselines.} We compare our method against NGLOD~\cite{nglod}, targeted at learning SDF representations with an octree-based feature volume. We also compare InstantNGP~\cite{InstantNGP}, NFFB~\cite{NFFB}, and NeuRBF~\cite{neuRBF}, where the hyper-parameters of those methods follow NeuRBF~\cite{neuRBF}.

\textbf{Dataset.} We use ten commonly used 3D models to benchmark the performances of 3D SDF reconstruction models. Those models are sampled by NeuRBF~\cite{neuRBF} and originated from public repositories~\cite{levoy2000digitalMichelangelo}.

\textbf{Results.} We present the quantitative result of 3D SDF models in~\Cref{tab-3d-recon} and learning curves are presented in the right of~\Cref{fig-comparison-curves}. Our model has achieved strong performance in the normal angular error (NAE) with fewer training parameters. Visualization results are provided in Supplementary~\Cref{supple-sec-sdf-reconstruction}.

\subsection{Novel view synthesis}
\label{sec-nerf-results}
\textbf{Model details.} The overall architecture for the view synthesis task is shown at the right of~\Cref{fig-model-ours}. We model the density $\boldsymbol{\sigma}$ via one \modelname:
\begin{subequations}
\begin{equation}
\boldsymbol{\sigma} = \phi_{sp}(\texttt{MulFAGrid}^{(1)}(\bx)),
\label{eq:MulFAGrid-model1}
\end{equation}
where $\phi_{sp}$ is the softplus activation~\cite{dvgo}:
\begin{equation}
\phi_{sp}(\bx) = \texttt{log}(1 + \texttt{exp}(\bx + \eta_b)).
\label{eq:softplus}
\end{equation}
The hyperparameter $\eta_b=1\times e^{-3}$ controls the empirical bias, which penalizes the number of non-zero elements in grid tensors to alleviate overfitting~\cite{dvgo}. We then model the color $\boldsymbol{c}$ via two terms considering position-dependent colors and view-dependent colors,
\begin{equation}
\boldsymbol{c}= \texttt{MLP}(\texttt{MulFAGrid}^{(2)}(\bx) \oplus \texttt{PosEmb}(\boldsymbol{d})),
\label{eq:MulFAGrid-model2}
\end{equation}
\end{subequations}
where $\texttt{MulFAGrid}^{(2)}$ denotes another \modelname, and $\texttt{PosEmb}$ refers to the positional embedding~\cite{dvgo, dvgov2}. After getting colors $(\boldsymbol{c}_i)_{1\leq i \leq N}$ and densities $(\boldsymbol{\sigma}_i)_{1\leq i \leq N}$ for $N$ segments, differentiable volumetric rendering~\cite{dvgo, nerf} is adopted to get the rendered pixel color.


\textbf{Baselines.} Besides previously mentioned baselines InstantNGP~\cite{InstantNGP}, NFFB~\cite{NFFB} and NeuRBF~\cite{neuRBF}, we also compare to other grid-based NeRF methods DVGOv2~\cite{dvgo, dvgov2} and Plenoxels~\cite{plenoxels}. We adopt regular grid-based models in this comparison, which is called ``Ours". We also provide a comparison with 3DGS~\cite{3DGS}, a recent very strong method in neural fields. Note that 3DGS~\cite{3DGS} requires external structure-from-motion to construct its point cloud, so when comparing to 3DGS~\cite{3DGS}, we also leverage the same point cloud (PC) as initialization, which means irregular grids are adopted in this case. Our method is called ``Ours (w/ PC)" in this case.

\textbf{Benchmarks.} We evaluate related methods on bounded NeRF benchmarks: LLFF~\cite{localLightFieldFusion} and SyntheticNeRF~\cite{nerf}. Besides, we also adopt two unbounded NeRF benchmarks: Tanks\&Temples~\cite{tanksAndTemples} and Mip-NeRF-360~\cite{mipnerf360}. Furthermore, we also conduct experiments on San Francisco Mission Bay (SFMB) provided by Block-NeRF~\cite{blocknerf}. Dataset details are presented in Supplementary~\Cref{supple-sec-nerf-dataset-details}.

\textbf{Results.} The quantitative comparison results on the five benchmarks are shown in~\Cref{tab-nerf-large-table}. Our model consistently improves over NeuRBF~\cite{neuRBF} for all the scenes, while~\modelname\space keeps a good balance between training speed and effectiveness. Furthermore, our method achieves competitive performance compared to the strong baseline of 3DGS~\cite{3DGS} when the point cloud initialization is derived by running the SfM initialization module of 3DGS~\cite{3DGS}. However, the rendering speed of 3DGS~\cite{3DGS} is still much faster than ours, partially due to their advanced CUDA implementation.

\subsection{Ablation studies}
\label{sec-ablation-study}
\begin{table}[t!]
\begin{center}
\resizebox{\columnwidth}{!}{%
\begin{tabular}{c|cc|cc}
\toprule
                     & \multicolumn{2}{c|}{2D image fitting}  & \multicolumn{2}{c}{3D SDF reconstruction}      \\ 
                     & PSNR$\uparrow$ & SSIM$\uparrow$ & IoU$\uparrow$ & NAE$\downarrow$ \\ \midrule
Learned kernel $\rightarrow$ interpolation & 48.12          & 0.9930         & 0.9991        & 5.24            \\
No normalization     & 56.01          & 0.9980         & 0.9995        & 4.81            \\
Fourier features $\rightarrow$ MSC             & 55.93          & 0.9978         & 0.9995        & 4.59            \\
Fourier features $\rightarrow$ SIREN           & 51.54          & 0.9958         & 0.9993        & 5.04            \\
Joint learning $\rightarrow$ decoupled            & 55.12          & 0.9980         & 0.9995        & 4.88            \\ \midrule
NeuRBF~\cite{neuRBF} & 54.84          & 0.9975         & 0.9995        & 4.93            \\
\rowcolor{gray} Ours full            & \textbf{56.19} & \textbf{0.9983}& \textbf{0.9995} & \textbf{4.51} \\ \bottomrule
\end{tabular}
}
\end{center}
\vspace{-10pt}
\caption{Ablation studies on 2D image fitting and 3D SDF reconstruction. We use the same validation dataset as that of NeuRBF~\cite{neuRBF}. Please refer to~\Cref{sec-ablation-study} for more details.}
\label{tab:ablation}
\vspace{-10pt}
\end{table}

We verify our design choices via ablation studies on the 2D image fitting dataset and the 3D SDF reconstruction dataset. Despite the mentioned changes, architectures and hyperparameters are kept the same. We train all models for 5k steps, and results are presented in~\Cref{tab:ablation}. Numbers of NeuRBF~\cite{neuRBF} and our full model are presented in the last two rows for reference.

We validate the effectiveness of adaptive learning in the first row, and we use a simple interpolation kernel to replace the learned kernel. The performance is significantly downgraded in this case, revealing the necessity of learning the kernel function $\varphi$. Then, we remove the normalization function (\Cref{eq-normalization}), and we find that the performance of the model slightly decreases. We replace the adopted Fourier features (\Cref{eq-fourier-features}) with the multi-frequency sinusoidal composition (MSC) proposed by NeuRBF~\cite{neuRBF}, where it shows slightly worse performance than our full model. In the fourth row, we replace the Fourier features with SIREN~\cite{SIREN} features, which demonstrates even worse performance. Lastly, we replace the joint optimization scheme of~\modelname\space with the decoupled learning method~\cite{neuRBF}, and we find that it has worse performance, which demonstrates the benefit of jointly learning the kernel parameters with grid features.

\section{Conclusion}
We have proposed the grid tangent kernel (GTK) theory, which grounded grid-based models and helped analyze their optimization and generalization performance. We conducted a fine-grained analysis inspired by GTK to explain the behaviors of grid-based models. Then, we proposed \modelname, a grid-based model for general neural field modeling. We proposed an adaptive learning scheme for \modelname\space by jointly optimizing the kernel parameters and grid features. \modelname\space balanced between ``underfitting" and ``overfitting" extremes and made more precise predictions. Meanwhile, \modelname\space supported both regular grids and irregular grids. Experimental results on 2D image regression, 3D SDF reconstruction, and novel view synthesis demonstrated that MulFAGrid achieved state-of-the-art performance compared to various grid-based models. We hope our theoretical findings can provide insight for designing better grid-based models.

\section{Acknowledgements}
Dr. Fenglei Fan would like to acknowledge that this paper was supported by the Direct Grant for Research from the Chinese University of Hong Kong and ITS/173/22FP from the Innovation and Technology Fund of Hong Kong. This work was also partly supported by NSFC (62222607) and the Shanghai Municipal Science and Technology Major Project (2021SHZDZX0102).

%% file: sec/X_suppl.tex
\clearpage
\setcounter{page}{1}
\maketitlesupplementary
\section{Proofs to theorems}
\label{supple-sec-proof-theorems}
One of our major contributions is to reveal the power of NTK theories~\cite{jacot2018NTK, fourierFFN, NTKarora2019fineGrained} to grid-based models. The main paper introduces several claims based on our introduced grid tangent kernel (GTK). We introduce proofs in this section and provide discussions and interpretations of claims in grid-based unbounded radiance fields.

\subsection{Settings}
\label{supSecNotations}

\begin{figure*}[t!]
\begin{center}
\includegraphics[width=1.0\textwidth]{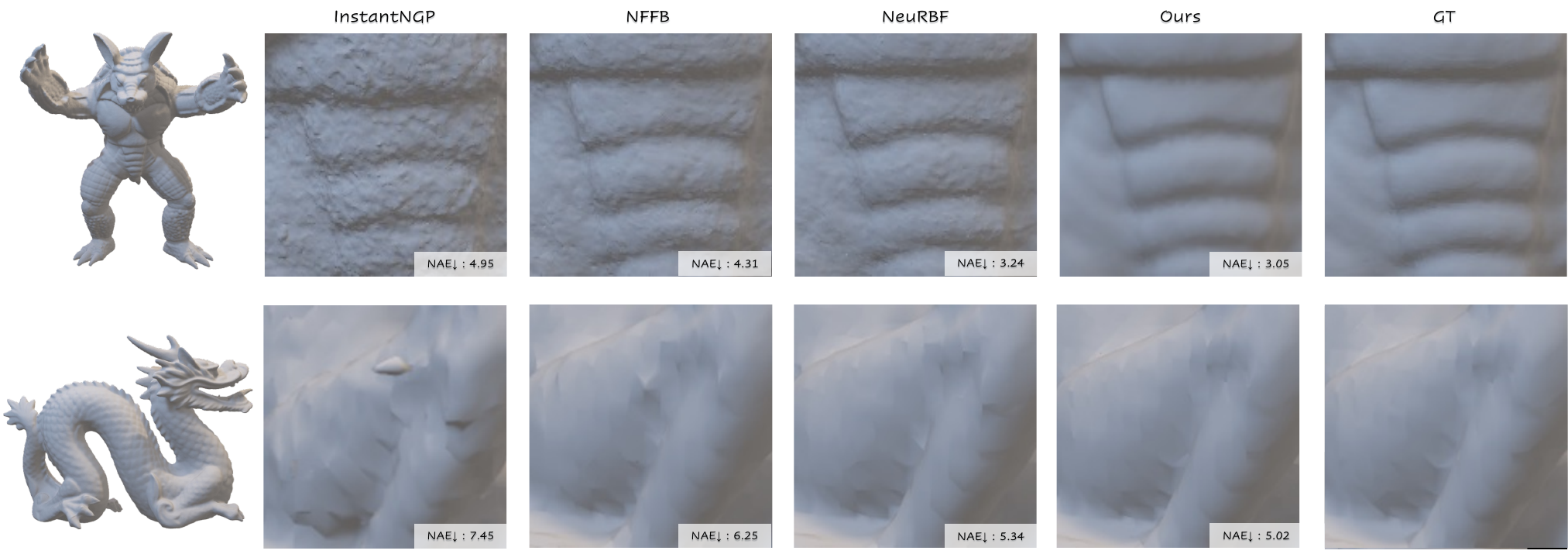}
\end{center}
\vspace{-10pt}
\caption{3D signed distance field (SDF) reconstruction results comparing to InstantNGP~\cite{InstantNGP}, NFFB~\cite{NFFB}, NeuRBF~\cite{neuRBF}. We show the reconstructed geometry of our approach at the leftmost column. We show the normal angular errors (NAE) in this figure.}
\label{fig-sdf-results}
\end{figure*}

We present our notations in the~\Cref{tab:notations}. Following previous works~\cite{fourierFFN, roberts2022principlesDeepLearning}, we build up the analysis framework for grid-based models in a supervised regression setting. For simplicity, we assume the model weights are initialized to zero tensors:
\begin{equation}
\bw(0) = \boldsymbol{0}.
\label{eqZeroInitialization}
\end{equation}

We use a regression loss to measure a grid-based model $g_{\bw}$ parameterized by $\bw$:
\begin{equation}
\mathcal{L}(\bw) = \frac{1}{2} \sum_{i=1}^n\left(\bY_i-g_{\bw}(\bX_i)\right)^2,
\label{eqLoss}
\end{equation}
and we assume $0\leq \bY_i \leq 1, 0 \leq g_{\bw} \leq 1$.
Model parameters evolve following gradient descent (GD), where the $r$-th weight $\bw_r$ can be written as:
\begin{equation}
\bw_r(t+1)-\bw_r(t)=-\eta_l \frac{\partial \mathcal{L}(\bw(t))}{\partial \bw_r}.
\label{eqGDDiscreteSingle}
\end{equation}
The vector form of this update rule is:
\begin{equation}
\operatorname{vec}(\bw(t+1)) = \operatorname{vec}(\bw(t)) - \eta_l \bZ(t)(\bO (t) - \bY_i),
\label{eqGDDiscreteVec}
\end{equation}
where $\bZ(t)$ is the gradient matrix at timestep $t$:
\begin{equation}
\bZ(t) = (\frac{\partial g\left(\bX_i, \bw(t)\right)}{\partial \bw_1}, \frac{\partial g\left(\bX_i, \bw(t)\right)}{\partial \bw_2}, ..., \frac{\partial g\left(\bX_i, \bw(t)\right)}{\partial \bw_r})^T.
\label{eqGradientMatrix}
\end{equation}

The continuous form of model dynamics can be described via gradient flow, which is an ODE~\cite{roberts2022principlesDeepLearning}:
\begin{equation}
\begin{aligned}
\frac{d \bw(t)}{d t}&=-\nabla \ml(\bw(t))\\&=-\sum_{i=1}^n\left(g\left(\bX_i, \bw(t)\right)-\bY_i\right) \frac{\partial g\left(\bX_i, \bw(t)\right)}{\partial \bw}.
\end{aligned}
\label{eqGradientFlow}
\end{equation}

\subsection{Grid-based models and their derivatives}
A grid model can be represented via a weighted average of features on the grid nodes, as shown in its definition:
\begin{equation}
g(\bX_i, \bw) \triangleq \sum_{i \in U(\bX_i)} \varphi\left(\bX_i, \Theta_i \right) \bw_i.
\label{eq-grid-model-repeat}
\end{equation}
For ease of mathematical analysis, we generalize the summation to all parameters instead of the surrounding index $U(\bX)$. Such a procedure can be achieved by setting the kernel $\varphi$ to zero at non-local indexes:
\begin{equation}
g(\bX_i, \bw) = \sum_{r=1}^m \varphi\left(\bX_i, \Theta_r \right) \bw_r.
\label{eqVoxelGrid}
\end{equation}
We further define the vector form of the nodal function as:
\begin{equation}
\varphi(\bX_i) = (\varphi(\bX_i, \Theta_1), \varphi(\bX_i, \Theta_2), ...,  \varphi(\bX_i, \Theta_m))^T.
\label{eqNodalVector}
\end{equation}
Grid-based models require that the kernel function $\varphi$ is only determined by the node index and the input data, and the kernel function is not changed during a concerned period of time. This constraint holds for many state-of-the-art grid-based models~\cite{InstantNGP, 3DGS, NFFB}. Although in some cases, such as adaptive learning~\cite{neuRBF}, the kernel function is optimized along with the feature vectors, the assumption is still a reasonable one by assuming that the kernel function is updated much slower than the grid features $\bw$. Therefore, during a short enough time span, the kernel function can be regarded as a static one.

The summarization of the kernel function is one (due to the normalization procedure~\Cref{eq-normalization}):
\begin{equation}
\sum_{r=1}^m \varphi(\bX_i, \Theta_r) = 1, \forall \bX_i.
\label{eqVoxelGridNodalFunctionSummarization}
\end{equation}





\begin{table}[t!]
\begin{center}
\resizebox{\columnwidth}{!}{%
\begin{tabular}{l|l}
\toprule \midrule
Variable                          & Definition                                                    \\ \midrule
$f$ & a function \\ \midrule
$g$ & a grid-based model \\ \midrule
$\bX_i$ & an input data \\ \midrule
$\bY_i$ & a label \\ \midrule
$n$ & the size of the dataset \\ \midrule
$m$ & the number of parameters in the model \\ \midrule
$d$ & the dimension of the output feature \\ \midrule
$\boldsymbol{A}$                  & a matrix                                                      \\ \midrule
$\boldsymbol{A}_{ij}$             & the $(i, j)$-th entry of $\boldsymbol{A}$                     \\ \midrule
$\|\cdot\|_2$                     & the Euclidean norm of a vector                                \\ \midrule
$\|\cdot\|_F$                     & the Frobenius norm of a matrix                                \\ \midrule
$\lambda_{\min }(\boldsymbol{A})$ & the minimum eigenvalue of a symmetric matrix $\boldsymbol{A}$ \\ \midrule
$\operatorname{vec}(\mathbf{A})$  & the vectorization of a matrix $\boldsymbol{A}$                \\ \midrule
$\boldsymbol{I}$ or $\boldsymbol{I}_n$                & the identity matrix with shape $n \times n$                   \\ \midrule
$\bw(t)$ & the weight matrix of shape $r\times d$ at timestep $t$ \\ \midrule
$\bw_r(t)$ & the $r$-th weight of the model at timestep $t$ \\ \midrule
$S=\{(\bX_i, \bY_i)\}_{i=1}^{n}$ & input-label samples \\ \midrule
$\bO(t) = g_{\bw}(\bX_i)$ = $g(\bX_i, \bw)$ & a model with weights $\bw$ and inputs $\bX_i$ \\ \midrule
$\bZ(t)$ & the gradient matrix at timestep $t$, see~\Cref{eqGradientMatrix} \\ \midrule
$\mg_g(t)$ & the GTK matrix of a grid-based model $g$ at timestep $t$ \\ \midrule
$\mathcal{L}$ & the loss\\ \midrule
$\eta_l$ & the learning rate\\ \midrule
$\gamma$ & the Fourier feature mapping\\\midrule
$\varphi$ & the nodal function\\\midrule
$\mathcal{F}$ & a function class \\ \midrule
$\bep$ & random variables from $\{-1, 1\}$ \\ \midrule
$B$ & an upper bound of weight change \\ \midrule
$B(\bw)$ & defined in~\Cref{eqBoundWeightB} \\ \midrule
$k_{o}$ & a coefficient used in~\Cref{eqZ0Bound} \\ \midrule
$\bt$ & defined in~\Cref{eqDefT} \\ \midrule
\bottomrule
\end{tabular}
}
\end{center}
\caption{Definitions of notations in this paper.}
\label{tab:notations}
\end{table}

\subsection{Proof of~\Cref{theorem-law}}

In~\Cref{theorem-law} of our paper, we show that the training dynamics of grid-based models are associated with the GTK:

\begin{proof}
The model parameters evolve according to the following differential equation:
\begin{equation}
\begin{aligned}
\frac{d g\left(\bX_i, \bw(t)\right)}{d t}=\frac{d \bw(t)}{dt} * \frac{\partial g\left(\bX_i, \bw(t)\right)}{\partial \bw}.
\end{aligned}
\end{equation}
Considering~\Cref{eqGradientFlow} and the definition of GTK, we have:
\begin{equation}
\frac{d g\left(\bX_i, \bw(t)\right)}{d t}=-\sum_{j=1}^n\left(g\left(\bX_j, \bw(t)\right)-\bY_j\right) [\mg_g(t)]_{i, j}.
\end{equation}
We can write it in a compact form, while considering the fact that $\bY$ is not changed during training:
\begin{equation}
\frac{d \bO(t)}{d t}=-\mg_g(t) \cdot(\bO(t)-\bY).
\label{eqUtEvolution}
\end{equation}
The discrete version of~\Cref{eqUtEvolution} with a learning rate $\eta_l$ can be written as:
\begin{equation}
\bO(t+1) - \bO(t) = \eta_l \mg_g(t) (\bO(t) - \bY).
\label{eqUpdateUKDiscrete}
\end{equation}
\end{proof}
\paragraph{Discussions.} Another equivalent definition of GTK given the gradient matrix $\bZ$ is: 
\begin{equation}
\mg_g(t) = \bZ(t)^T\bZ(t).
\label{eqGTKDefByGradientMatrix}
\end{equation}
This theorem shows that GTK connects
the error term $\bO(t)-\bY$ to the changing rate of the output. Therefore, this theorem can be used to analyze the training behaviors of grid-based models. We further show in~\Cref{theorem-gtk-unchanged} that GTK keeps constant during training, and therefore, standard kernel regression methods~\cite{roberts2022principlesDeepLearning} can be applied to analyze behaviors of grid-based models.

\subsection{Proof of~\Cref{theorem-gtk-unchanged}}
\begin{proof}
According to the previous analysis, the kernel function $\varphi$ remains constant during training. Therefore, according to~\Cref{eqGradientMatrix} and~\Cref{eqVoxelGrid}, the gradient matrix can be written as:
\begin{equation}
\bZ(t) = \varphi(\bX_i).
\label{eqGradientMatrixDetailed}
\end{equation}
Therefore, the gradient matrix remains constant during training:
\begin{equation}
\bZ(t) = \bZ(0) = \bZ.
\label{eqZconstant}
\end{equation}

According to~\Cref{eqGTKDefByGradientMatrix}, we can conclude that the GTK is not changed across training. Therefore, we have:
\begin{equation}
\mg_{g}(t) = \mg_{g}(0),
\label{eqGTKconstant}
\end{equation}
\end{proof}

\paragraph{Discussions.} This theorem shows that the GTK is unchanged during training for grid-based models. Therefore, GTK is a powerful tool for understanding these grid-based models' training and generalization properties. We can call these grid-based models as \textit{quasi-linear} models because although the model is not linear regarding the input data $\bX_i$, the model is linear regarding the weights. Different from NTK~\cite{jacot2018NTK}, which is constant only when the network width is infinite, the property of GTK is not asymptotic, which means that grid-based models might be better understood than conventional neural networks (MLPs).

\subsection{Proof of~\Cref{theorem-gtk-generalization}}
\begin{proof}
The major technique uses the empirical Rademacher complexity to bound the population loss according to the following theorem from~\cite{NTKarora2019fineGrained}. We first recap the definition of population loss and empirical loss:
\begin{equation}
\mathcal{L}_\mathcal{D}\left(t\right)=\mathbb{E}_{(\bX_i, \bY_i) \sim \mathcal{D}}\left[\mathcal{L}\left(f\left(\bX_i, {\bw(t)}\right), \bY_i\right)\right],
\end{equation}
\begin{equation}
\mathcal{L}_S\left(t\right) = \frac{1}{n} \sum_{i=1}^{n} \mathcal{L}\left(f\left(\bX_i, {\bw(t)}\right), \bY_i\right).
\end{equation}
Then, we introduce a useful theorem from~\cite{NTKarora2019fineGrained}.

\begin{theorem}
(from \textbf{Theorem B.1} of~\cite{NTKarora2019fineGrained}) Given a set of $n$ samples $S$, the empirical Rademacher complexity of a function class $\mathcal{F}$ (mapping from $\mathbb{R}^d$ to $\mathbb{R}$) is defined as:
\begin{equation}
\mathcal{R}_S(\mathcal{F})=\frac{1}{n} \mathbb{E}_{\bep \in\{ \pm 1\}^n}\left[\sup _{f \in \mathcal{F}} \sum_{i=1}^n \bep_i f\left(\bX_i\right)\right],
\end{equation}
where $\bep$ contains i.i.d random variables drawn from a uniform Rademacher distribution in $\{-1, 1\}$. Given a bounded loss function $\mathcal{L}(\cdot, \cdot)$, which is 1-Lipschitz in the first argument. Then with probability at least $1-\delta_p$ over sample $S$ of size $n$:
\begin{equation}
\sup _{f \in \mathcal{F}}\left\{\mathcal{L}_{\mathcal{D}}(f)-\mathcal{L}_S(f)\right\} \leq 2 \mathcal{R}_S(\mathcal{F})+3 c \sqrt{\frac{\log (2 / \delta_p)}{2 n}}.
\end{equation}
\label{TheoremRademacher}
\end{theorem}
Given a bound $B>0$ (we will calculate $B$ in our case later), we consider a bounded function of grid-based models:
\begin{subequations}
\begin{equation}
\mathcal{F}_{B}^{\bw(0)}=\left\{g_{\bw}:\right.\left.B(\bw)\right\},
\end{equation}
\begin{equation}
B(\bw) \triangleq \|\bw-\bw(0)\|_F \leq B.
\label{eqBoundWeightB}
\end{equation}
\label{EqBoundedClass}
\end{subequations}
We calculate the empirical Rademacher complexity as follows:
\begin{equation}
\begin{aligned}
\mathcal{R}_S(\mathcal{F}_{B}^{\bw(0)})&=\frac{1}{n} \mathbb{E}_{\bep \in\{ \pm 1\}^n}\left[\sup _{f \in \mathcal{F}_{B}^{\bw(0)}} \sum_{i=1}^n \bep_i g\left(\bX_i\right)\right]\\ &= \frac{1}{n} \mathbb{E}_{\bep \in\{ \pm 1\}^n}\left[\sup _{B(\bw)} \sum_{i=1}^n \bep_i \sum_{r=1}^m \varphi(\bX_i, \Theta_r) \bw_r \right],
\end{aligned}
\end{equation}
where $\varphi(\bX_i, \Theta_i)$ is the kernel function, and here we levearage~\Cref{eqVoxelGrid}.

Considering~\Cref{eqZconstant}, we can write the above equation as:
\begin{equation}
\begin{aligned}
\mathcal{R}_S(\mathcal{F}_{B}^{\bw(0)})&= \frac{1}{n} \mathbb{E}_{\bep \in\{ \pm 1\}^n}\left[\sup _{B(\bw)} \operatorname{vec}(\bw)^T\bZ\bep\right]\\&= \frac{1}{n} \mathbb{E}_{\bep \in\{ \pm 1\}^n}\left[\sup _{B(\bw)} \operatorname{vec}(\bw)^T\bZ(0)\bep\right]\\&= \frac{1}{n} \mathbb{E}_{\bep \in\{ \pm 1\}^n}\left[\sup _{B(\bw)} \operatorname{vec}(\bw-\bw(0))^T\bZ(0)\bep\right]\\&\leq\frac{1}{n}\mathbb{E}_{\bep \in\{ \pm 1\}^n}\left[B \cdot\|\bZ(0) \bep\|_2\right]\\&\leq\frac{B}{n} \sqrt{\underset{\bep \sim\{ \pm 1\}^n}{\mathbb{E}}\left[\|\bZ(0) \bep\|_2^2\right]} \\&=\frac{B}{n}\|\bZ(0)\|_F.
\end{aligned}
\label{eqRsF1}
\end{equation}
We first bound $\|\bZ(0)\|_F$:
\begin{equation}
\begin{aligned}
\|\bZ(0)\|_F^2&=\sum_{r=1}^{m} \sum_{i=1}^{n} \varphi^2(\bX_i, \Theta_r)\\ &= \sum_{i=1}^{n} \sum_{r=1}^{m} \varphi^2(\bX_i, \Theta_r)\\ &\leq k_{o}n.
\end{aligned}
\label{eqZ0Bound}
\end{equation}
The design and weights of the kernel function affect the constant $k_{o}$. Since it is a common practice in NTK theories~\cite{NTKarora2019fineGrained, jacot2018NTK} that we scale the output (and therefore the gradient) of the network by a constant, we set $k_{o}=\frac{1}{2}$ to make the resulting generalization bound consistent with that in the NTK theory~\cite{NTKarora2019fineGrained}. The exact value of $k_{o}$ does not affect the conclusions of our analysis, and it's safe to set $k_{o}=\frac{1}{2}$. We now have the following:
\begin{equation}
\mathcal{R}_S(\mathcal{F}_{B}^{\bw(0)}) \leq \frac{1}{\sqrt{2n}} B.
\label{eqRsF2}
\end{equation}

Then we prove~\Cref{eqBoundWeightB} holds and calculates the upper bound $B$ in the equation. We start from~\Cref{eqUpdateUKDiscrete} and apply~\Cref{eqGTKconstant}:
\begin{equation}
\bO(k+1) - \bO(k) = \eta_l \mg_g(\bO(k) - \bY).
\end{equation}
Recursively applying the above equation, we can derive the following:
\begin{equation}
\begin{aligned}
\bO(k) - \bY &= - (\bi - \eta_l \mg_g)^k (\bO(0) - \bY) \\
&= - (\bi - \eta_l \mg_g)^k \bY.
\end{aligned}
\end{equation}
Here, we use the assumption that the model weights are all set to zero stated in~\Cref{eqZeroInitialization}, and we use the definition of the model predictions in~\Cref{eqVoxelGrid}. We introduce a polynomial of $\mg_g$ as:
\begin{equation}
\bt \triangleq \sum_{i=0}^{k-1}\eta_l(\bi - \eta_l \mg_g)^i
\label{eqDefT}
\end{equation}
Then we plug the above result including~\Cref{eqZconstant} into~\Cref{eqGDDiscreteVec}:
\begin{equation}
\begin{aligned}
\|\bw(k) - \bw(0)\|_F &= \sqrt{\|\operatorname{vec}\left(\bw(k)\right) - \operatorname{vec}\left(\bw(0)\right)\|^2_2}\\ &= \sqrt{\left\|\sum_{i=0}^{k-1}\eta_l\bZ(\bi - \eta_l \mg_g)^i \bY \right\|_2^2}\\&=\sqrt{\|\bZ \bt \bY\|_2^2} \\ &= \sqrt{\bY^T\bt^T \bZ^T \bZ \bt \bY} \\ &= \sqrt{\bY^T\bt \bZ^T \bZ \bt \bY} \\ &= \sqrt{\bY^T\bt \mg_g \bt \bY},
\end{aligned}
\label{eqWKdist1}
\end{equation}
where we use~\Cref{eqGTKDefByGradientMatrix} and we consider the fact that $\bt$ is a symmetric matrix.
Decompose the matrix of $\mg_g$ as follows:
\begin{equation}
\mg_g = \sum_{i=1}^n \lambda_i \bv_i \bv_i^{\top}.
\end{equation}
Since $\bt$ is a polynomial of $\mg_g$, its eigenvectors are the same as $\mg_g$, and we have:
\begin{equation}
\begin{aligned}
\bt&=\sum_{i=1}^n \eta_l \sum_{j=0}^{k-1}\left(1-\eta_l \lambda_i\right)^j \bv_i \bv_i^{\top}\\ &=\sum_{i=1}^n \frac{1-\left(1-\eta_l \lambda_i\right)^k}{\lambda_i} \bv_i \bv_i^{\top}.
\end{aligned}
\end{equation}
Therefore, we have:
\begin{equation}
\begin{aligned}
\bt \mg_g \bt&=\sum_{i=1}^n\left(\frac{1-\left(1-\eta_l \lambda_i\right)^k}{\lambda_i}\right)^2 \lambda_i \bv_i \bv_i^{\top}\\ &\preceq \sum_{i=1}^n \frac{1}{\lambda_i} \bv_i \bv_i^{\top}\\&=\left(\mg_g\right)^{-1}.
\end{aligned}
\end{equation}
Plug this into~\Cref{eqWKdist1}, we have:
\begin{equation}
\|\bw(k) - \bw(0)\|_F \leq \sqrt{\bY^{\top}\mg_g^{-1}\bY}.
\end{equation}
Here we have proved~\Cref{eqBoundWeightB}.

Set $B=\bY^{\top}\mg_g^{-1} \bY$ and plug into~\Cref{eqRsF2}, we have:
\begin{equation}
\mathcal{R}_S(\mathcal{F}_{B}^{\bw(0)}) \leq \sqrt{\frac{\bY^{\top}\mg_g^{-1}\bY}{2n}}.
\end{equation}
Then we are ready to apply~\Cref{TheoremRademacher}:

\begin{equation}
\begin{aligned}
LHS&=\sup_{f \in \mathcal{F}_{B}^{\bw(0)}}\left\{\mathcal{L}_{\mathcal{D}}(f)-\mathcal{L}_S(f)\right\}\\ &\leq \sqrt{\frac{2\bY^{\top}\mg_g^{-1}\bY}{n}}+3\sqrt{\frac{\log (\frac{2}{\delta_p})}{2 n}}\\ &=\sqrt{\frac{2\bY^{\top}\mg_g^{-1}\bY}{n}} + O\left(\sqrt{\frac{\log \frac{2}{\delta_p}}{n}}\right).
\end{aligned}
\end{equation}

Bound $\mathcal{L}_S(f)$ as follows (considering~\Cref{eqLoss}):
\begin{equation}
\begin{aligned}
\mathcal{L}_S(f) & =\frac{1}{n} \sum_{i=1}^n\left[\mathcal{L}\left(\bO_i(k), \bY_i\right)-\mathcal{L}\left(\bY_i, \bY_i\right)\right] \\
& \leq \frac{1}{n} \sum_{i=1}^n\left|\bO_i(k)-\bY_i\right| \\
& \leq \frac{1}{\sqrt{n}}\|\bO(k)-\bY\|_2 \\
& =\sqrt{\frac{2 \mathcal{L}(\bw(k))}{n}} \\
& \leq \frac{1}{\sqrt{n}},
\end{aligned}
\end{equation}
where we use the fact that the loss $\mathcal{L}(\bw(k))\leq \frac{1}{2}$.
Therefore, we have proved~\Cref{theorem-gtk-generalization}.
\end{proof}
\paragraph{Discussions.} The key insight behind this proof procedure is that the generalization gap is strongly associated with weight change during training. Therefore, if we can narrow down the required weight change across training (e.g., adding more inductive bias or setting proper initialization), we will have a model that generalizes better. 

Another insight is that the generalization gap is associated with the dominating term $\Delta = \bY^{\top}\mg_g^{-1}\bY$, which is a quadratic form of $\mg_g^{-1}$. We may use this knowledge to motivate designs of future grid-based models better. 

Also, labels $\bY$ will affect the generalization ability of grid-based models. This could explain why pose initialization accuracy and point cloud initializations greatly matter to radiance field reconstruction~\cite{nerf++, blocknerf, 3DGS}.

\subsection{NeRF experimental details}
\label{supple-sec-nerf-exp-details}
In unbounded scenes, one must warp the scene into normalized device coordinates (NDC~\cite{nerf,mip_nerf}) before feeding the coordinates into neural networks. Following Mip-NeRF 360~\cite{mipnerf360} and DVGOv2~\cite{dvgov2}, we use a two-layer parameterization to model near and far objects separately. Formally, for a sampled point $\bp$ in the ray $\br$ where its real-world coordinate $\bX_w^{\bp}$ is transferred to the normalized one:
\begin{equation}
\bX_n^{\bp} = \begin{cases} \bX_w^{\bp}, & \|\bX_w^{\bp}\|_{\infty} \leq 1, \\ \left(1+b-\frac{b}{\|\bX_w^{\bp}\|_\infty}\right)\frac{\bX_w^{\bp}}{\|\bX_w^{\bp}\|_\infty}, & \|\bX_w^{\bp}\|_{\infty}>1,\end{cases}
\label{eq:ndc-unbounded}
\end{equation}
where $b$ is a hyperparameter of the background length. We set $b=0.2$ in all experiments.

The Fourier feature length $l$ is set to five in all the experiments. The size of the voxels in grid-based models for NeRF experiments is set to $320^3$ ($N_x=N_y=N_z=320$). The post-activation bias is set to $\eta_b =1\times e^{-3}$. We use $\eta_{\mathcal{F}}=0.5$ to balance two losses. We use the Adam optimizer~\cite{kingma2014adam} with a batch size of $8192$ rays to optimize the representation for $40k$ iterations. We use a constant learning rate $\eta_{l}=1 \times e^{-3}$ without learning rate decay. Our implementations are based on Pytorch. The speed test is conducted on a single NVIDIA 3090Ti GPU card and averaged numbers across three runs are reported to avoid random noises. The other setup of the speed test follows previous works~\cite{dvgov2,neuRBF}. More details are in the supplementary.

\section{Additional NeRF dataset details}
\label{supple-sec-nerf-dataset-details}
We produce additional NeRF dataset details on unbounded datasets here.
\textbf{Tanks\&Temples~\cite{tanksAndTemples}.} We show experimental results on four large-scale scenes provided by~\cite{tanksAndTemples}. All the scenes are hand-held 360-degree captures, and camera poses are estimated by COLMAP~\cite{schoenberger2016sfm}. We use the same dataset split as DVGOv2~\cite{dvgov2}.

\textbf{Mip-NeRF-360~\cite{mipnerf360}.} A dataset of seven scenes was published. Each scene contains a challenging central object in the background with rich details. Camera poses are derived via COLMAP~\cite{schoenberger2016sfm}. Comparison experiments follow the dataset split of previous work~\cite{mipnerf360, dvgov2}.

\textbf{San Francisco Mission Bay (SFMB)~\cite{blocknerf}.} This is a street scene dataset released by Block-NeRF~\cite{blocknerf}. The images are captured in San Francisco's Mission Bay District. Twelve cameras on a vehicle record images from different angles. Collected images are divided into \texttt{train} and \texttt{test} splits. We generate a virtual driving camera sequence, including rotating and forwarding poses, in the \texttt{render} split. We do not compare to the full version of Block-NeRF~\cite{blocknerf}, which is not open-sourced. However, we re-implement their block division technique for baselines and ours: we train separate models for different blocks, and renderings are generated via the block composition~\cite{blocknerf}. The number, sizes, and positions of blocks are the same for all compared methods. Please refer to the supplementary for more details. On SFMB~\cite{blocknerf}, we use one block to measure the number of parameters and the training time.

\section{3D SDF reconstruction visualizations}
\label{supple-sec-sdf-reconstruction}
We provide visualizations of SDF estimation in~\Cref{fig-sdf-results}. From this visualization, we can find that our results can recover more fine-grained details compared to other methods. Moreover, our results often produce more smooth surfaces compared to other baselines.